\documentclass[12pt]{article}
\usepackage{graphicx}
\usepackage[utf8]{inputenc}
\usepackage{lscape}
\usepackage{float}
\usepackage{mathtools}
\usepackage{longtable}
\usepackage{amsfonts}
\usepackage{amsmath}
\usepackage{mathabx}
\usepackage{graphicx}
\usepackage{bm}
\usepackage{xcolor}
\usepackage{appendix}
\usepackage{relsize}
\usepackage{outlines}
\usepackage{blkarray}
\usepackage[ruled,vlined]{algorithm2e}
\usepackage{algpseudocode}
\usepackage[margin=1in]{geometry}
\usepackage{amsthm}
\usepackage{thmtools}
\usepackage{caption}
\usepackage{subcaption}
\usepackage[normalem]{ulem}
\useunder{\uline}{\ul}{}

\DeclareMathOperator*{\argmin}{arg\,min}
\declaretheoremstyle[
headfont=\normalfont\bfseries,
  notefont=\mdseries, notebraces={(}{)},
  bodyfont=\normalfont,
  spaceabove = 0.5cm, 
  spacebelow = 0.5cm
]{mystyle}
\declaretheorem[style=mystyle]{example}

\title{RieszBoost: Gradient Boosting for Riesz Regression}
\author{Kaitlyn J. Lee\\
Division of Biostatistics\\ 
UC Berkeley  \and Alejandro Schuler\\Division of Biostatistics\\ UC Berkeley}

\begin{document}

\maketitle
\begin{abstract}
    Answering causal questions often involves estimating linear functionals of conditional expectations, such as the average treatment effect or the effect of a longitudinal modified treatment policy. By the Riesz representation theorem, these functionals can be expressed as the expected product of the conditional expectation of the outcome and the Riesz representer, a key component in doubly robust estimation methods. Traditionally, the Riesz representer is estimated indirectly by deriving its explicit analytical form, estimating its components, and substituting these estimates into the known form (e.g., the inverse propensity score). However, deriving or estimating the analytical form can be challenging, and substitution methods are often sensitive to practical positivity violations, leading to higher variance and wider confidence intervals. In this paper, we propose a novel gradient boosting algorithm to directly estimate the Riesz representer without requiring its explicit analytical form. This method is particularly suited for tabular data, offering a flexible, nonparametric, and computationally efficient alternative to existing methods for Riesz regression. Through simulation studies, we demonstrate that our algorithm performs on par with or better than indirect estimation techniques across a range of functionals, providing a user-friendly and robust solution for estimating causal quantities.
\end{abstract}

\section{Introduction}

Researchers are often interested in estimands of the form
$$\Psi(\mathbb{P}_0)=\mathbb{E}[m(O,\mu_0)],$$
where $O=(Y,W)$ are observations drawn from a distribution $\mathbb{P}_0$ (i.e., expectations are with respect to $\mathbb{P}_0$; we write $\mathbb{E}[\cdot]$ in place of $\mathbb{E}_0[\cdot]$ for brevity) and $\mu_0(W) = \mathbb{E}[Y|W]$. If $\mu \mapsto \mathbb{E}[m(O,\mu)] $ is linear and continuous for $\mu \in L^2$, then by the Riesz representation theorem we can re-write our parameter as
$$
\mathbb{E}[m(O, \mu_0)] = \mathbb{E}[\alpha_0(W)\mu_0(W)],
$$

\noindent where $\alpha_0(W)$ is referred to as the \textit{Riesz representer}. The Riesz representation theorem is nothing other than the infinite-dimensional analogue of the fact that any fixed linear mapping from $d$-dimensional vectors $v$ to numbers can be written as a dot product between the input $v$ and some other fixed vector.\footnote{Let $f$ be a linear function from vectors $v \in \mathbb R^d$ to numbers $\mathbb R$. Any $v$ can be written in terms of the bases $e_1 \dots e_d$ of the vector space as $v = \sum_{j=1}^d v_j e_j$. Since $f$ is linear, $f(v) = f\left(\sum_{j=1}^d v_j e_j\right) = \sum_{j=1}^d v_j f(e_j) = v \cdot a$ with $a_j = f(e_j)$ and thus $a$ is the Riesz representer of the function $f$. In the general case we can think of a function $g(w)$ as an infinite vector: one entry per input $w$, i.e. the argument is the equivalent of the ``index'' $j$ for finite vectors. In this case $f$ maps functions to numbers and an appropriate inner product is $\mathbb{E}[\alpha(W) g(W)] = \int \alpha(w) g(w) \, d \mathbb{P}(w)$ which generalizes the familiar dot product (an infinite sum of products of the elements of the two ``vectors'' at the ``index'' $w$). For a fuller explanation, consult \cite{kreyszig_introductory_1989}.} 

\begin{example}[Average Treatment Effect (ATE)]
    \label{ex:ate}
      Consider the average treatment effect, a common parameter of interest in causal inference \cite{robins_estimation_1994, rubin_neymans_2015}. Let $W=(A, X)$, where $A$ is a binary treatment of interest and $X$ are confounders. Under causal identification assumptions, the causal ATE can be written as 
    $$ \Psi(\mathbb{P}_0) = \mathbb{E}[\mathbb{E}[Y|A=1, X]-\mathbb{E}[Y|A=0, X]].$$
    For the ATE, $\mu: (a,x) \mapsto \mathbb{E}[Y|A=a,X=x]$, and the linear functional of interest is given by 
    \begin{equation}
    \label{eq:ate-functional}
        m(O,\mu) = \mu(1, X) - \mu(0, X).
    \end{equation}
    Using standard conditioning arguments, it can also be shown that
    $$ \Psi(\mathbb{P}_0) = \mathbb E\left[\left(\frac{A}{\mathbb{P}(A=1|X)} - \frac{1-A}{\mathbb{P}(A=0|X)}\right) \mu_0(A,X)\right]$$
    which motivates the well-known inverse propensity score-weighted estimator. From this we can see that $\alpha_0(A,X) = \frac{A}{\mathbb{P}(A=1|X)} - \frac{1-A}{\mathbb{P}(A=0|X)}$, i.e. the inverse propensity weights.
\end{example}

Knowing the Riesz representer typically motivates weighting estimators because by conditioning we have $ \Psi(\mathbb{P}_0) = \mathbb{E}[\alpha_0(W)\mu_0(W)] = \mathbb{E}[\alpha_0(W) Y] \approx \frac{1}{n} \sum_i^n \alpha_0(W_i) Y_i $. Thus, the values $\alpha_0(W_i)$ represent appropriate ``weights'' on the outcomes $Y_i$ that can be used to estimate the parameter of interest. But weighting estimators are typically more variable than alternatives, which motivates the use of ``efficient'' or ``doubly robust'' estimators. These estimators are desirable because they the smallest variance among all regular asymptotically linear estimators and are often still consistent under some kinds of model misspecification. 

Efficient estimators of such estimands still rely upon Riesz representation because they usually require estimates of the outcome regression $\mu_0$ and the Riesz representer $\alpha_0$. This is because such estimation strategies rely upon characterizing the semi-parametric efficiency bound of such estimands by the efficient influence function (EIF) (see, e.g., \cite{robins_estimation_1994,
bickel_efficient_1998, van_der_laan_unified_2003, tsiatis_semiparametric_2006, robins_quadratic_2009}). As shown in \cite{chernozhukov_automatic_2022}, the EIF (at $\mathbb{P}_0$) for a parameter $\Psi: \mathbb{P}_0 \mapsto \mathbb{E}[m(O, \mu_0)] = \mathbb{E}[\alpha_0(W)\mu_0(W)]$ involves the Riesz representer and is given by
\begin{equation}
\label{gen_eif}
\phi_0(O) = m(O, \mu_0) -  \Psi(\mathbb{P}_0) + \alpha_0(W)(Y-\mu_0(W)).
\end{equation}

\renewcommand{\thmcontinues}[1]{ATE, continued}
\begin{example}[continues=ex:ate]
The EIF for the ATE is given by 
    \begin{equation}
    \label{ate_eif}
        \phi_0(O) = \mu_0(1,X)-\mu_0(0,X) -  \Psi(\mathbb{P}_0) + \left(\frac{A}{\pi_0(X)} - \frac{1-A}{1-\pi_0(X)}\right) \big(Y-\mu_0(W)\big),
    \end{equation}
    \noindent where $\pi_0(X) = \mathbb{P}(A=1|X)$ is the true propensity score function and $\alpha_0(A,X) = \frac{A}{\pi_0(X)} - \frac{1-A}{1-\pi_0(X)}$ is the Riesz representer for the ATE.
\end{example}

Many estimators have been constructed using the EIF as a foundation. For example, the efficient estimating equations (EEE) estimator for this class of estimators is formed by by first estimating  $\mu_0$ and $\alpha_0$, substituting these into the EIF in place of the true functions, setting the empirical average of the result equal to 0, and solving for $ \Psi(\mathbb{P}_0)$ \cite{robins_estimation_1994, van_der_laan_unified_2003}. When applied to the ATE, this approach results in the famous augmented inverse propensity-weighted (AIPW) estimator. Targeted minimum loss-based estimation (TMLE) also begins by estimating $\mu_0$. Then, an estimate of $\alpha_0$ is used in a ``targeting'' step, which adjusts the initial $\mu_0$ estimate $\hat \mu_0$.  Finally, the updated version of $\hat \mu_0$ is plugged into $\mathbb{E}[m(O,\hat \mu_0)]$  \cite{laan_targeted_2006}. In the TMLE literature, the ``clever covariate'' used in targeting for many one-step updates is precisely the Riesz representer evaluated at the observed data.

Typically, researchers estimate $\alpha_0$ indirectly by first deriving the form of $\alpha_0$ explicitly, estimating the relevant functions, and then plugging the estimates in. For example, when estimating the ATE, one usually estimates the propensity score $\pi_0(X)$ and then plugs it into the known form $\alpha_0(A,X) = \frac{A}{\pi_0(X)} - \frac{1-A}{1-\pi_0(X)}$. However, the analytical form of $\alpha_0$ can be difficult to derive for more complex causal estimands. Even if it can be derived, estimating the relevant components may be difficult - for example, plug-in estimation of the Riesz representer of some generalized average treatment effects, which allow for longitudinal data structures and continuous treatment variables as introduced in \cite{susmann_longitudinal_2024}, involves numerical integration for each observation. Additionally, the form of $\alpha_0$ often involves estimating quantities that appear in the denominator of some terms. For example, for the ATE, $\pi_0(X)$ appears in the denominator of the Riesz representer. In finite samples, if there are certain populations that have very small or very large probabilities of being treated, then such substitution estimators can produce highly variable estimates.

Chernozhukov et al. recently developed Riesz regression, a method of directly estimating $\alpha_0$ from the data without needing to derive the analytical form of the Riesz representer itself \cite{chernozhukov_automatic_2024}. Their work provides results for the Riesz loss, demonstrating that its minimizer under the true data-generating process is the true Riesz representer for the target parameter. In particular, the authors propose algorithms leveraging neural networks and modified random forests to optimize the Riesz loss \cite{chernozhukov_riesznet_2022}. Riesz regression is a technique within the broader balancing weights literature, where inverse probability weights are estimated using the method of moments to align covariate distributions across treatment groups of interest \cite{ben-michael_balancing_2021}. Other balancing weight methods include entropy balancing weights \cite{hainmueller_entropy_2012}, inverse probability tilting \cite{graham_inverse_2012}, and stable balancing weights \cite{zubizarreta_stable_2015} - regression adjustment can also be rewritten as a balancing weights problem \cite{chattopadhyay_implied_2022}.

In this paper, we develop an algorithm to minimize the Riesz loss using gradient boosting. Gradient boosting is a supervised ensemble learning algorithm in which many weak learners are added together to form the final prediction \cite{friedman_greedy_2001}. Gradient boosting a general method: it works with different loss functions and different weak learners. When the weak learners used are trees, the algorithm is referred to as gradient boosted trees (GBTs). GBT has been shown to consistently outperform other supervised machine learning algorithms in simulations and online competitions \cite{caruana_empirical_2006, chen_xgboost_2016, ke_lightgbm_2017}. Additionally, research has shown that, in general when working with tabular data, gradient boosted trees outperform neural networks, while also being easier and less costly to train \cite{shwartz-ziv_tabular_2022, mcelfresh_when_2024, borisov_deep_2024}. Thus, we argue that implementing Riesz regression with gradient boosted trees provides a more user-friendly approach to Riesz regression than the algorithms currently found in the literature, without sacrificing performance or generality.

The rest of the paper is organized as follows. Section \ref{methods} describes the RieszBoost algorithm in more detail and the special considerations researchers must attend to when considering gradients of the Riesz loss. Section \ref{sims} presents simulation study results. Finally, Section \ref{discuss} contains a discussion about our algorithm and results.

\section{Methods}\label{methods}

We consider settings in which the observed data $O \sim \mathbb{P}_0$ consists of an outcome of interest $Y$ and regressors $W$. If one is interested in estimating causal parameters, we often have $W = (A, X)$, where $A$ is the treatment of interest and $X$ is a vector of pre-treatment confounders. Let $\mu_0(W) = \mathbb{E}[Y|W]$ be the true outcome regression under $ \mathbb{P}_0$. Say we observe $n$ i.i.d. copies of $O$ such that $O_i = (Y_i, W_i)$ for $i = 1, \dots, n$. We use boldface to denote the vector or matrix of all $n$ observations of a given random variable, e.g., $\mathbf Y = [Y_1, \dots Y_n]^\top$. When we use notation like $f(\mathbf Z),$ we mean an elementwise application $[f(Z_1), \dots f(Z_n)]^\top.$ Similarly, we abuse $z \in \mathbf Z$ to mean that $z$ is one of the values taken by the random variable $Z$ in the observed data.

\subsection{Riesz Regression}
\label{riesz_reg}
We seek to minimize the Riesz loss, as developed in \cite{chernozhukov_automatic_2024}. In the paper, Chernozhukov et al. show that one can write $\alpha_0$ as the minimizer of the Riesz loss function:
\begin{align*}
\alpha_0 &= \argmin_{\alpha} \mathbb{E}[(\alpha_0(W)-\alpha(W))^2]\\
&= \argmin_{\alpha}  \mathbb{E}[-2\alpha_0(W)\alpha(W)]  + \mathbb{E}[\alpha(W)^2]\\
&= \argmin_{\alpha}  \mathbb{E}[-2m(O, \alpha) + \alpha(W)^2],
\end{align*}
where the last equality follows from $\mathbb{E}[m(O, \alpha)]=\mathbb{E}[\alpha_0(W)\alpha(W)]$ for functions in $L^2$ (Riesz representation applied to $\alpha$). 

We call $l(O, \alpha) = -2m(O, \alpha) + \alpha(W)^2$ the \textit{Riesz loss} for a parameter $\Psi$. Estimators that minimize this loss in expectation are termed ``Riesz regressions," as they are the solution to the population-level least-squares regression problem with ``target'' $\alpha_0(W)$ and regressors $W$. In practice, the expected loss $\mathbb{E}[l(O_i, \alpha)]$ is not available so we instead minimize the empirical Riesz loss $\frac{1}{n}\sum_{i=1}^n l(O_i, \alpha)$. 

Note that the loss only depends on the function $m(O, \mu)$ and \textit{not} on how $\alpha_0$ depends on the unknown distribution $\mathbb{P}_0$. For example, in estimating an average treatment effect, we could theoretically use this loss to learn the inverse propensity weights (the Riesz representer) without even knowing that these weights should be inverse propensity scores. In this way, Riesz regression is closely related to the balancing weights literature.

\renewcommand{\thmcontinues}[1]{ATE, continued}
\begin{example}[continues=ex:ate]
  Given the functional in equation \ref{eq:ate-functional}, the empirical Riesz loss for the ATE is 
  \begin{equation}
  \label{eq:ate-loss}
  L_n(\alpha)
  = \frac{1}{n} \sum_{i=1}^n -2(\alpha(1, X_i) - \alpha(0, X_i)) + \alpha(A_i, X_i)^2.
  \end{equation}
  We see in this example that the empirical loss depends on the values of $\alpha$ at observed data $(A_i, X_i)$ for $i = 1, \dots, n$, as well as the values at counterfactual data $(0, X_i)$ for $i$ such that $A_i = 1$ and $(1, X_i)$ for $i$ such that $A_i = 0$.
\end{example}

\paragraph{}Before moving on, we make an observation that will be important later: as we see in the example above, the empirical Riesz loss often depends not only on the values of $\alpha$ at observed data but also on the values at ``pseudo-data.'' When estimating causal parameters, these pseudo-data are often the counterfactual observations of interest. This differs from other regression loss functions. For example, consider the mean squared error (MSE) loss for outcome regression:
$$l(O, \mu) = (Y-\mu(W))^2.$$
When minimizing the empirical MSE $L_n = \frac{1}{n}\sum_{i=1}^n l(O_i, \mu)$ over candidate functions $\mu$, one only needs to consider the values of $\mu$ at observed points $O_i$ for $i \in \{1, \dots, n\}$.

\subsection{Gradient Boosting}
\label{gbt}

In this paper, we employ gradient boosting to directly learn the Riesz representer by minimizing the Riesz loss. We describe gradient boosting in terms of gradient descent in function space, as outlined in \cite{friedman_greedy_2001}.

To understand gradient boosting, it helps to first consider gradient descent in a parametric setting. Suppose we aim to estimate a parameter $\beta_0$, where $\beta_0$ minimizes a given loss function over candidates $\beta$. Starting with an initial guess, gradient descent updates the estimate iteratively. At each step, the gradient of the loss function with respect to $\beta$ is calculated at the current estimate. This estimated gradient provides the direction of steepest descent, indicating how we should adjust $\beta$ to reduce the loss. We then update our estimate by taking a small step in the direction of the estimated negative gradient. Repeating this process brings us progressively closer to the true minimizer of the loss, $\beta_0$.

Gradient boosting follows a similar principle but operates in function space. The goal is to minimize an empirical loss function $\alpha \mapsto \frac{1}{n}\sum_{i=1}^n l(O_i, \alpha(W_i))$ over candidate functions $\alpha$. Here, let $L(\alpha)=\mathbb{E}[l(O,\alpha(W))]$ be the population loss and $L_n(\alpha) = \frac{1}{n}\sum_{i=1}^n l(O_i, \alpha(W_i))$ be its empirical counterpart.

Similar to gradient descent, we use the gradient of the loss to guide updates. However, in contrast to parametric gradient descent, we take the gradient of the loss with respect to candidate functions $\alpha$. For a function $f: \mathcal X \to \mathbb R$, we use $\nabla_{x_0} f \in \mathcal X$ to mean the gradient of $f(x)$ with respect to $x$ evaluated at a point $x_0$. In the parametric setting, $\beta_0 \in \mathbb{R}^p$ is finite-dimensional, and the gradient $\nabla_{\beta}L$ represents the typical derivative of $L(\beta)$ with respect to $\beta$. We can denote the $j$th partial derivative $\nabla_{\beta} L(j)=\frac{\partial{L}}{\partial{\beta_j}}$.

It is helpful to think of the function of interest $\alpha(w)$ as an infinite-dimensional parameter vector, with one parameter for every possible value of the regressor $w$ (think of $w$ as the ``index'' for the vector $\alpha$). At each boosting step, the goal is to update the estimated function $\alpha(w)$ by taking a small step in the direction of the gradient $\nabla_\alpha L(w)$. Like $\alpha(w)$, this gradient can also be viewed as an infinite-dimensional vector indexed by $w$, providing the direction of steepest descent in function space at every point.

To update an initial estimate $\alpha_m(w) \rightarrow \alpha_{m+1}(w)$, we would ideally apply an update like
$$\alpha_{m+1}(w) = \alpha_m(w) - \lambda \nabla_{\alpha_m}L(w),$$
where $\lambda$ (the ``learning rate'') is typically a small number so that we only take small steps in the direction of the gradient. However, we do not know the true function $\nabla_{\alpha_m}L(w)$; instead, we must estimate this gradient from data. 

To make things more concrete, we will walk through how to employ gradient boosting for estimating the outcome regression. Once again, consider the MSE loss:
$$l(O, \mu) = (Y-\mu(W))^2.$$ One way to estimate the true gradient $\nabla_{\mu}L(w)$ is by using its empirical analog $\nabla_{\mu}L_n(w)$. At each step, we are able to calculate the empirical gradient:

\begin{equation}
\label{eq:mse-grad}
\begin{aligned}
\nabla_{\mu}L_n (w)
&= \nabla_\mu \left[ \frac{1}{n} \sum_{i=1}^n (Y_i-\mu(W_i))^2 \right] (w) \\
&= \left[ \frac{1}{n} \sum_{i=1}^n \left[\nabla_\mu (Y_i-\mu(W_i))^2 \right](w) \right]  \\
&= \frac{1}{n} \sum_{i=1}^n -2(Y_i-\mu(W_i))1(w=W_i).
\end{aligned}
\end{equation}

This estimate, however, is noisy and can only be non-zero at observed values $w \in \mathbf W$. For all other values, $\nabla_{\mu}L_n(w)=0$. Therefore, if we were to naively perform an update using
$$\mu_{m+1}(w) = \mu_m(w) - \lambda \nabla_{\mu_m}L_n(w),$$
our estimate for $\mu_{m+1}(w)$ could only ever be updated at observed $w \in \mathbf W$. Our estimates for unobserved values of $w$ would always remain unchanged and our resulting prediction function would be useless for out-of-sample prediction.

A natural way of improving this noisy estimate of the gradient is to perform a regression to smooth it. We can regress the initial estimates $\nabla_{\mu}L_n(w)$ onto $w\in \mathbf W$ to approximate the population level gradient $\nabla_{\mu}L(w)$. This yields an estimated gradient $\widehat{\nabla_\mu L}(w)$, which is defined for all values of $w$, not just those observed in the data. In gradient boosting, a ``weak learner'' (e.g., a regression tree) is typically used to  regress $\nabla_{\mu_m}L_n(w)$ onto $w \in \mathbf W$, yielding predictions $f_{m+1}(w) = -\widehat{\nabla_{\mu_m} L}(w)$ at each iteration $m$. We then apply the update
$$\mu_{m+1}(w) = \mu_m(w) + \lambda f_{m+1}(w)$$
and repeat this process $M$ times. Starting with an initial estimate $f_0(w) = 0$, the final estimate is
$$\widehat \mu(w) = \sum_{m=1}^M \lambda f_m(w).$$

\paragraph{Implementation} For squared error loss and most other loss functions used in gradient boosting, the gradient is only nonzero at the values of $w$ observed in the data and therefore must only be computed at those values. Moreover, the value of the gradient at a value $W_i$ depends only on the current estimate $\hat Y_i = \hat\mu(W_i)$ and data $(Y_i, W_i)$ corresponding to the same observation $i$. For squared error loss, for example, $\nabla_{\hat\mu} L_n(W_i) = -2(Y_i - \hat Y_i)$ follows from equation \ref{eq:mse-grad}.\footnote{This holds under the simplifying assumption that there are no ties in the observed values of $W$. In practice, the ``generalized residuals'' $R_i, R_j$ corresponding to two data points with identical regressors $W_i = W_j$ may be different and are allowed to ``average out'' through the regression even though both technically represent the empirical gradient at the same point.} Therefore, the computation of the gradient is typically performed row-wise, applying some ``residual function'' $R_i = r(Y_i, \hat Y_i)$ to each observation $O_i$ to produce the empirical (negative) gradient $R_i = -\nabla_{\hat\mu} L_n(W_i)$ at that point. In the case of squared error, we have that $r: (Y,\hat Y) \mapsto 2n^{-1}(Y-\hat Y)$, often written as $Y - \hat Y$ since the constant factors may be absorbed into the learning rate. 

A typical implementation is shown in Algorithm \ref{algo:gb}. Besides the indicated hyperparameters, the user should indicate the algorithm used to fit the weak learners (generally decision trees of a given depth, etc.). The residual function $r$ is often set automatically from the type of outcome (e.g., MSE gradient for continuous, Bernoulli log-likelihood gradient for binary, etc.).

\begin{algorithm}[H]
\label{algo:gb}
\caption{Gradient Boosting}
\SetKwProg{Fn}{Function}{:}{}
\SetKwFunction{fit}{fit}
\SetKwFunction{predict}{predict}
\SetKwFunction{append}{append}

\KwIn{
    Input data: $\mathbf{W}, \mathbf{Y}$; \\
    Hyperparameters: $\lambda, M$; \\
    Regression algorithm: \fit; \\
    Residual function: $r$
}
\KwOut{Estimated function: \predict}
\

Initialize $\mathbf{\hat{Y}} \gets 0$, $\mathbf{R} \gets 0$, and $f \gets []$\;

\For{$m \gets 1$ \textbf{to} $M$}{
    Compute residuals: $\mathbf{R} \gets r(\mathbf{Y}, \mathbf{\hat{Y}})$\;
    Fit model: $f_m \gets \fit(\mathbf{R}, \mathbf{W})$\;
    Update weak learner list: $f.\append(f_m)$\;
    Update predictions: $\mathbf{\hat{Y}} \gets \mathbf{\hat{Y}} + \lambda f_m(\mathbf{W})$\;
}

\Fn{\predict{$w$}}{
    Initialize prediction: $\hat{y} \gets 0$\;
    \For{$m \gets 1$ \textbf{to} $M$}{
        Update prediction: $\hat{y} \gets \hat{y} + \lambda f_m(w)$\;
    }
    \Return $\hat{y}$
}

\Return \predict\;

\end{algorithm}

\subsection{RieszBoost}
\label{rieszboost}

Our contribution in this paper is a method for implementing Riesz regression using gradient boosting. As described in Section \ref{gbt}, this requires estimating the population-level gradient function $\nabla_{\alpha}L$ at each boosting step. Conceptually, nothing needs to be modified in the abstract gradient boosting framework as previously described to enable Riesz regression, apart from substituting the regression loss with the Riesz loss corresponding to the parameter of interest. However, in practice, this replacement results in subtle modifications to the standard gradient boosting algorithm, which we discuss here. To mitigate these changes, we propose a data-augmentation ``trick'' that minimizes implementation differences. 

As mentioned in Section \ref{riesz_reg}, a key difference between standard regression losses and Riesz losses is that Riesz losses in many cases evaluate the candidate function at multiple points of interest. In particular, for many causal parameters (where $W = (A,X)$), we find that the gradient $\nabla_{\alpha}L(a,x)$  depends not only on the observed data $\mathbf W = (\mathbf A, \mathbf X)$ but also some relevant counterfactual data points $\{\mathbf a, \mathbf X\}$ for particular values of $a$. This makes standard boosting implementations fail because they only compute and track values of the empirical gradient at the observed data. 

\renewcommand{\thmcontinues}[1]{ATE, continued}
\begin{example}[continues=ex:ate]
    Differentiating the loss from equation \ref{eq:ate-loss} and ignoring constants (which are absorbed into the learning rate) gives
    \begin{equation}
    \begin{aligned}
    \label{eq:ate-grad-1}
        \nabla_{\alpha}L_n(a,x) 
        &= \sum_{i=1}^n 1\big((a,x) = (A_i,X_i)\big)\alpha(a,x) - 1(x = X_i)\big(1(a=1)-1(a=0)\big)
        .
    \end{aligned}
    \end{equation}
\end{example}

\paragraph{}The simplest way to keep track of the gradient and Riesz representer at values besides those observed in the data is to augment the data with any necessary ``pseudo-data'' (e.g., counterfactual observations) at which we expect the empirical gradient to be nonzero. 

\renewcommand{\thmcontinues}[1]{ATE, continued}
\begin{example}[continues=ex:ate]
    To evaluate the gradient in equation \ref{eq:ate-grad-1}, we can construct the ``predictor matrix'' with $2n$ rows:
    $$
    \tilde{\mathbf W} = 
    \begin{blockarray}{cc}
    \tilde{A} & \tilde{X} \\
    \begin{block}{[cc]}
    \mathbf A & \mathbf X  \\
    \mathbf{1}-\mathbf{A} & \mathbf X  \\
    \end{block}
    \end{blockarray}
    .
    $$
    The rows of this matrix represent all of the values $w=(a,x)$ at which the gradient is non-zero (for at least one square-integrable $\alpha$). In this case, we have added rows to the observed data corresponding to counterfactual treatments for each individual. 
    
    The value of the gradient at these points depends on the current prediction at the predictor point but also on the specific values of the original and counterfactual treatment. This is analogous to the fact that, in standard regression settings, the gradient depends on the predicted value $\hat Y$ but also the observed outcome $Y$. Therefore, we define the ``target matrix'' with $2n$ rows:
    $$
    \tilde {\mathbf Z} = 
    \begin{blockarray}{cc}
    \tilde{A} & A^{\degree} \\
    \begin{block}{[cc]}
    \mathbf A & \mathbf A \\
    \mathbf{1}-\mathbf{A} & \mathbf A \\
    \end{block}
    \end{blockarray}
    ,
    $$
    \noindent with each row corresponding to a row in the predictor matrix $\tilde {\mathbf W}$. 
    
    For all rows of the predictor matrix $(\tilde A_j, \tilde X_j) \in \tilde{\mathbf W}$, we know that there is some $i$ for which $\tilde X_j = X_i \in \mathbf X$ by construction. This eliminates the indicator function for $X$ in eq. \ref{eq:ate-grad-1}. Assuming rows of $\mathbf W$ are unique, the value of the empirical gradient evaluated at a such a point in the prediction matrix is therefore
    \begin{equation}
    \begin{aligned}
    \label{eq:ate-grad-2}
        \nabla_{\alpha}L_n(\tilde A_j,\tilde X_j) 
        &=1\big(\tilde A_j = A^\degree_j\big)\alpha(\tilde A_j,\tilde X_j) - \big(1(\tilde A_j =1)-1(\tilde A_j=0)\big)
        .
    \end{aligned}
    \end{equation}
    
    At the moment, this is slightly imprecise because we have not articulated how $\nabla_\alpha L_n$ has access to $A^\degree$ nor how it matches the appropriate rows. This will be formalized shortly. Nonetheless, the main idea should be clear in this example. With these pieces in place, we can define a row-wise ``residual'' function where, for each row $j = 1, \dots, 2n$, we evaluate
    $$r(\tilde Z_j, \hat Z_j) = -1(\tilde{A}_j=A^\degree_j)\hat Z_j  + (2\tilde{A}_j - 1)$$
    where we have defined $\hat Z_j = \hat\alpha(\tilde 
    A_j, \tilde X_j)$ as the estimated $\hat \alpha$ applied to row $j$ in predictor matrix $\tilde {\mathbf{W}}$ and $\tilde{Z}_j = (\tilde A_j,A^\degree_j)$ as row $j$ of the target matrix $\tilde{\mathbf{Z}}$. 
\end{example}

\paragraph{}Our proposed boosting algorithm for Riesz regression is fully described in Algorithm \ref{algo:rieszboost}. Using the data augmentation trick to define the predictor and target matrices minimizes the differences with Algorithm \ref{algo:gb}. The only differences are 1) the use of the predictor dataset as the input data and 2) the fact that the target matrix is now multidimensional.
This algorithm easily accommodates advancements and optimizations for standard gradient boosting: for example, to implement stochastic Riesz boosting, we simply execute the fitting step with a random subsample of the residuals and predictor matrix \cite{friedman_stochastic_2002}. We show only the most basic version here to keep the presentation accessible.

\begin{algorithm}[H]
\label{algo:rieszboost}
\caption{RieszBoost}
\SetKwProg{Fn}{Function}{:}{}
\SetKwFunction{fit}{fit}
\SetKwFunction{predict}{predict}
\SetKwFunction{append}{append}

\KwIn{
    Predictor and target matrices: $\mathbf{\tilde W}$, $\mathbf{\tilde Z}$; \\
    Hyperparameters: $\lambda, M$; \\
    Regression algorithm: \fit; \\
    Residual function: $r$
}
\KwOut{Estimated function: \predict}
\

Initialize $\mathbf{\hat{Z}} \gets 0$, $\mathbf{R} \gets 0$, and $f \gets []$\;

\For{$m \gets 1$ \textbf{to} $M$}{
    Compute residuals: $\mathbf{R} \gets r(\mathbf{\tilde Z}, \mathbf{\hat Z})$\;
    Fit model: $f_m \gets \fit(\mathbf{R}, \mathbf{\tilde W})$\;
    Update weak learner list: $f.\append(f_m)$\;
    Update predictions: $\mathbf{\hat{Z}} \gets \mathbf{\hat{Z}} + \lambda f_m(\mathbf{\tilde W})$\;
}

\Fn{\predict{$w$}}{
    Initialize prediction: $\hat{z} \gets 0$\;
    \For{$m \gets 1$ \textbf{to} $M$}{
        Update prediction: $\hat{z} \gets \hat{z} + \lambda f_m(w)$\;
    }
    \Return $\hat{z}$
}

\Return \predict\;

\end{algorithm}

In our setup, each unique parameter implies a different formulation of the predictor matrix $\mathbf{\tilde W}$, target matrix $\tilde{\mathbf{Z}}$, and the residual function $r$. To construct $\mathbf{\tilde W}$, we augment the original regressors $\mathbf{W}$ by adding rows corresponding to the inputs to $\alpha$ that result in a non-zero empirical gradient. For many causal parameters of interest, the additional rows will correspond to the counterfactual values of treatment of interest and covariates for each individual. Formally we can write the (random) matrix $\tilde {\mathbf W} = \{w: \nabla_\alpha L_n \ne 0 \text{ for some } \alpha \in \mathcal L_2\}$ as a function of all the original regressors $\mathbf W$ and the functional $m$: $\tilde {\mathbf W}(\mathbf W, m)$. In all the examples we consider we have that $\tilde {\mathbf W}$ is finite and that each element $\tilde { W}_j$ depends on the matrix $\mathbf W$ through only a single row $ W_{i(j)}$. Formally, we can write $\tilde {\mathbf W}(\mathbf W, m) = \sqcup_i \tilde W(W_i,m)$ with $\tilde W$ a mapping from points to sets.
In this case, elements in $\tilde W(W_i,m)$ depend only on $W_i$, i.e. there is (reverse) mapping from indices $j$ of $\tilde {\mathbf W}$ to indices $i$ of $\mathbf W$.

For any row $\tilde{W}_j$ in $\tilde{\mathbf{W}}$, we attempt to write the empirical gradient at $\tilde{W}_j$ in the form 
$\nabla_\alpha L_n(\tilde{W}_j) 
= r\big(
\tilde{z}(\tilde{W_j}, W_{i(j)}), \alpha(\tilde{W}_j)
\big)$. 
Here $\tilde Z_j = \tilde{z}(\tilde{W_j}, W_{i(j)})$ captures the part of the empirical gradient that does not depend on the candidate function $\alpha$ whereas $\hat Z_j = \alpha(\tilde W_j)$ captures the part that does. This mapping allows us to construct the target matrix $\tilde{\mathbf{Z}}$ row-wise by applying $\tilde{z}$ to each row of the predictor matrix (pulling in the corresponding row $W_{i(j)}$ of the original regressors). For many causal parameters, $\tilde{\mathbf{Z}}$ will include a column corresponding to the counterfactual exposures of interest and a column corresponding to the observed exposure. These techniques work in great generality (at least for all the examples we have considered).

Although we use the notation $\tilde Z$ for a row of the target matrix and $\hat Z$ for the estimated value of the Riesz representer at a point, $\hat Z$ is \textit{not} an estimate of $\tilde Z$, and in fact these objects generally have different dimensions. Our notation is meant to draw an analogy to the usual residual function that compares a target and prediction.

We demonstrate our construction in an additional example below and two others (average treatment effect among the treated and local average shift effect) in Appendix \ref{appendix:ex_params}.

\begin{example}[Average Shift Effect (ASE)]
\label{ex:ase}
 We now present the average shift effect (ASE) as a second example of how to implement Riesz boosting, in this case for a causal effect of a \textit{continuous} treatment. This parameter was described in \cite{diaz_population_2012, diaz_nonparametric_2023}.

    Consider again data $O=(Y,A,X)$, where $A$ is a now a continuous exposure of interest. We are interested in estimating the average shift effect (ASE), the expected outcome under an additive increase in treatment (covariate values held constant) relative to the average observed outcome:
    $$ \Psi(\mathbb{P}_0) = \mathbb{E}[\mathbb{E}[Y|A+\delta, X]-\mathbb{E}[Y|A, X]].$$
    Under standard identification assumptions, this parameter captures the causal effect of increasing treatment by $\delta$ units across the population.
    
    The linear functional of interest is given by $m(O,\mu) = \mu(A+\delta, X) - \mu(A, X)$. For this parameter, we have $\alpha_0(A,X) = \frac{p_{A|X}(A-\delta,X)}{p_{A|X}(A,X)} - 1$, where $p_{A|X}(A,X)$ is the true conditional density of treatment given confounders. We do not need to know this to do Riesz regression, but we include it for completeness. For completeness, we also include the EIF:
    $$\phi_0(O) = \mu_0(A+\delta, X) - \mu_0(A,X) -  \Psi(\mathbb{P}_0) + \left( \frac{p_{A|X}(A-\delta,X)}{p_{A|X}(A,X)} - 1\right)\big(Y-\mu_0(W)\big).$$
    
    Given the form of $m(O,\mu)$ shown above, the empirical Riesz loss is
    $$L_n(\alpha) = \frac{1}{n} \sum_{i=1}^n -2(\alpha(A_i+\delta, X_i) - \alpha(A_i, X_i)) + \alpha(A_i, X_i)^2.$$
    And, taking derivatives, we compute the empirical gradient:
    \begin{align*}
    \nabla_{\alpha}L_n(a,x) &= \sum_{i=1}^n 1\big((a,x) = (A_i,X_i)\big)\alpha(a,x) - 1(x =X_i)\big(1(a=A_i+\delta)-1(a=A_i)\big)
   \end{align*}
   To evaluate the empirical gradient, we must evaluate candidate functions $\alpha$ at points $(a,x)$ and $(a+\delta, x)$ for $a \in \mathbf{A}$ and $x \in \mathbf{X}$. Therefore, we can construct the predictor matrix with $2n$ rows:
   $$
    \tilde{\mathbf W} = 
    \begin{blockarray}{cc}
    \tilde{A} & \tilde{X} \\
    \begin{block}{[cc]}
    \mathbf A & \mathbf X  \\
    \mathbf{A} + \delta & \mathbf X  \\
    \end{block}
    \end{blockarray}.
    $$
    
    \noindent We also see that the empirical gradient at each row in the predictor matrix depends on the original and counterfactual treatment values. Therefore, we can construct the target matrix:
    
    $$
    \tilde {\mathbf Z} = 
    \begin{blockarray}{cc}
    \tilde{A} & A^\degree \\
    \begin{block}{[cc]}
    \mathbf A & \mathbf A  \\
    \mathbf A+\delta & \mathbf A \\
    \end{block}
    \end{blockarray}.
    $$

    Finally, we can define the residual function where, for each $j = 1, \dots, 2n$, we evaluate

    $$r(\tilde{Z}_j, \hat{Z}_j) = -1(\tilde A_j=A^\degree_j)\hat{Z}_j + \big(1(\tilde{A}_j \neq A^\degree_j) - 1(\tilde{A}_j = A_j^\degree)\big),$$

    \noindent where $\tilde{Z}_j = \hat{\alpha}(\tilde A_j, \tilde X_j)$ and $\hat{Z}_j = (\tilde A_j, A^\degree_j)$. We use the predictor matrix, target matrix, and residual function in Algorithm \ref{algo:rieszboost} to implement boosted Riesz regression for the ASE.
\end{example}

\paragraph{Usage.} Estimated Riesz representers from RieszBoost may be used in any downstream inference framework (e.g., TMLE, double machine learning (DML), EEE) as applicable. In almost all cases, it is recommended to use cross-fitting \cite{zheng_asymptotic_2010, chernozhukov_doubledebiased_2018}: the Riesz representer should be fit in one sample and used to generate predictions for a different sample (see references and Appendix \ref{appendix:est} for further details). RieszBoost accommodates this without modification. 

\paragraph{Tuning.} It is good practice to tune machine learning estimators, and RieszBoost is no different. We recommend constructing the \textit{validation-set} empirical Riesz loss $L_{\Acute n}: \alpha \mapsto \frac{1}{\Acute n}\sum_{i \in \Acute{\mathbf O}}^{\Acute n} \alpha(W_i)^2 - 2m(O_i, \alpha)$ where $\mathbf{\Acute O}$ represents a set of $\Acute n$ independent draws of the data. The validation Riesz loss can be used to choose between representers estimated with different algorithms or hyperparamters (e.g., boosting iterations $M$, learning rate $\lambda$, choice of base learner)   \cite{singh_kernel_2024, craven_smoothing_1978}. This concept generalizes to \textit{cross}-validation using the Riesz loss, which makes more efficient use of the data. If using cross-fitting, cross-validation can be nested inside the cross-fitting folds, although in practice one may not lose much by using validation folds as estimation folds as well (see ``cross-validated cross-estimation'' in \cite{wager_high-dimensional_2016}).

An important advantage of RieszBoost over alternatives is that tuning is relatively straightforward. We recommend using early stopping in the number of iterations $M$ \cite{zhang_boosting_2005} and a simple grid search in the learning rate $\lambda$. If regression trees are used as base learners, the most important tuning parameter is usually tree depth - this can also be tuned with a small grid.

\section{Simulation Studies}\label{sims}

To demonstrate the utility of our gradient boosting method, we perform simulation studies and compare the performance of our estimator with that of procedures that estimate the Riesz representer indirectly via nuisance parameter estimation (e.g., estimating the propensity score and then inverting). We use two different data generating processes to allow for estimation of the ATE, average treatment effect among the treated (ATT), ASE, and local average shift effect (LASE), an example of a generalized ATT. The details for boosting for the ATE and ASE were provided in Examples \ref{ex:ate} and \ref{ex:ase}, respectively. Details of the loss and gradient derivations ATT and LASE can be found in Appendix \ref{appendix:ex_params}. To estimate the parameters and variance, we implement efficient estimating equations (EEE) estimators, which are semi-parametrically efficient and doubly-robust \cite{robins_estimation_1994}. Details on the estimation strategies used can be found in Appendices \ref{appendix:est} and \ref{appendix:ex_params}. We evaluate all methods in terms of estimation error for the Riesz regression and estimation error for the final target parameter (e.g., ATE).

\subsection{Binary treatment}

To simulate data with a binary treatment variable, we use the following data generating process (DGP): 
\begin{align*}
X &\sim \text{Uniform}(0,1)\\
A|X &\sim \text{Binomial}(p = \text{logit}(-0.02X -X^2 + 4 \text{log}(X+0.3) +1.5))\\
Y|A,X &\sim \text{Normal}(\mu = 5X + 9XA + 5 *\text{sin}(X\pi) + 25(A-2), \sigma^2 = 1) \\
\end{align*}
The estimands of interest are the ATE and ATT. Under the DGP, the true values of the parameters are given by $\psi^{ATE} = 29.502$ and $\psi^{ATT} = 30.786$. 

At each iteration, we estimate the Riesz representer for the ATE and ATT both directly using our gradient boosting algorithm for Riesz regression and indirectly by first estimating the propensity score function (using GBT for classification) and then substituting this estimate into the known form of the Riesz representer for each parameter. Details of these estimation techniques for the ATT can be found in Appendix \ref{appendix:est/rr_att}.  We also estimate the outcome regression using GBT. Finally, we compute estimates $\widehat{\psi}^{ATE}$ and $\widehat{\psi}^{ATT}$ using EEE estimators for the two parameters that leverage the estimated outcome regression and Riesz representer (estimated either directly or indirectly).

We draw 1,000 observations from the DGP in each of 500 simulations. We use 500 observations as our training data set and the other 500 serve as our estimation data set (this is a simple version of cross-fitting). We use cross validation to train each gradient boosting algorithm over a set grid of hyperparameters on the training data. Specifically, we consider learning rate values $\lambda \in \{0.001, 0.01, 0.1, 0.25\},$ number of boosting iterations $M\in\{10, 30, 50, 75, 100, 150, 200\}$, and we use trees of maximum depth $\in \{3,5,7\}$ as our base learners. The hyperparameter grid used was the same for the RieszBoost estimator, GBT for propensity score learning, and GBT for outcome regression learning.

\subsubsection{Results for ATE and ATT}
To illustrate the estimation strategies visually, we plot the estimated Riesz representer function for the ATE over values of $X$ for the treated ($A=1$) and control ($A=0$) groups for one data set. The red dotted line represents the true function $\frac{A}{\pi_0(X)} - \frac{1-A}{1-\pi_0(X)}$. The blue line represents the RieszBoost estimates, and the green line represents the indirect estimates substituting the propensity score estimates.

\begin{figure}[h!]
\centering
\begin{subfigure}{.5\textwidth}
  \centering
  \includegraphics[width=\linewidth]{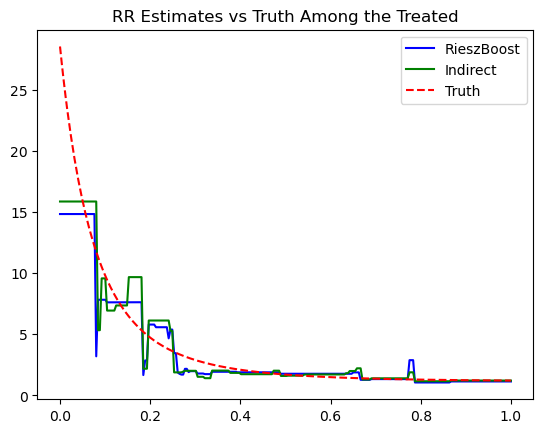}
  \label{fig:ate_treated}
\end{subfigure}%
\begin{subfigure}{.5\textwidth}
  \centering
  \includegraphics[width=\linewidth]{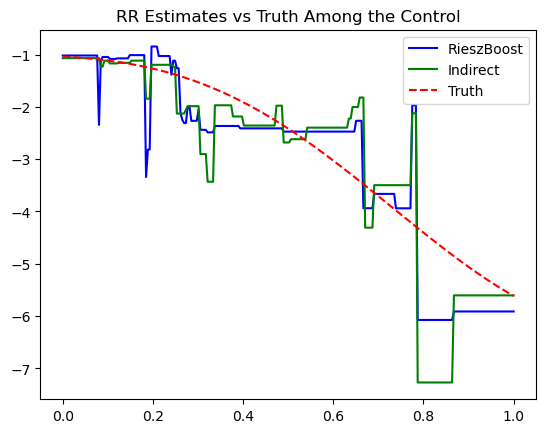}
  \label{fig:sub2}
\end{subfigure}
\label{fig:ate_control}
\end{figure}

 In Table
\ref{tab:rr_rmse_binary}, we compare the performance with regards to estimating the Riesz representer as a function in terms of root mean squared error (RMSE) and mean absolute error (MAE) over the 500 simulations. The direct RieszBoost estimator outperforms the indirect method for both the ATE and ATT, with lower average  RMSE and MAE. 

Tables \ref{tab:ate_results} and \ref{tab:att_results} present our main results for estimating the ATE and ATT, respectively. The improved performance in estimating the Riesz representer shown in Table \ref{tab:rr_rmse_binary} translates into better estimation of the causal parameters of interest. While both methods produce unbiased estimates for the ATE and ATT, for the ATE, RieszBoost estimates yield tighter confidence intervals and better coverage for the true value at $\alpha = 0.05$ for the ATE. For the ATT, the results for RieszBoost and the indirect method are fairly comparable, with RieszBoost resulting in smaller RMSE and slightly better coverage. 

\begin{table}[h!]
\centering
\caption{Results for Estimating $\alpha_0^{ATE}$ and $\alpha_0^{ATT}$}
\label{tab:rr_rmse_binary}
\begin{tabular}{ccccc}
                                      &                                    &                                   &                                    &                                   \\ \cline{2-5} 
\multicolumn{1}{c|}{}                 & \multicolumn{2}{c|}{\textbf{ATE}}                                      & \multicolumn{2}{c|}{\textbf{ATT}}                                      \\ \hline
\multicolumn{1}{|c|}{\textbf{Method}} & \multicolumn{1}{c|}{\textbf{RMSE}} & \multicolumn{1}{c|}{\textbf{MAE}} & \multicolumn{1}{c|}{\textbf{RMSE}} & \multicolumn{1}{c|}{\textbf{MAE}} \\ \hline
\multicolumn{1}{|c|}{RieszBoost}      & \multicolumn{1}{c|}{0.920}         & \multicolumn{1}{c|}{0.347}         & \multicolumn{1}{c|}{0.435}         & \multicolumn{1}{c|}{0.185}         \\
\multicolumn{1}{|c|}{Indirect}        & \multicolumn{1}{c|}{1.402}         & \multicolumn{1}{c|}{0.577}         & \multicolumn{1}{c|}{0.636}         & \multicolumn{1}{c|}{0.278}         \\ \hline
\end{tabular}
\end{table}

\begin{table}[h!]
\centering
\caption{ATE Simulation Results}
\label{tab:ate_results}
\begin{tabular}{|c|c|c|c|c|c|}
\hline
\textbf{Method}     & \textbf{Avg. Estimate} & \textbf{Avg. Est. SD} & \textbf{RMSE} & \textbf{Empirical SD} & \textbf{Coverage (95\%)} \\ \hline
RieszBoost & 29.522                 & 0.175                 & 0.187         & 0.186                 & 0.940                    \\
Indirect   & 29.539                 & 0.176                 & 0.260         & 0.257                 & 0.902                    \\ \hline
\end{tabular}
\end{table}

\begin{table}[!h]
\caption{ATT Simulation Results}
\centering
\label{tab:att_results}
\begin{tabular}{|c|c|c|c|c|c|}
\hline
\textbf{Method}     & \textbf{Avg. Estimate} & \textbf{Avg. Est. SD} & \textbf{RMSE} & \textbf{Empirical SD} & \textbf{Coverage (95\%)} \\ \hline
RieszBoost & 30.786                 & 0.173                 & 0.177         & 0.177                 & 0.950                    \\
Indirect   & 30.793                 & 0.175                 & 0.191         & 0.191                 & 0.942                    \\ \hline
\end{tabular}
\end{table}

\subsection{Continuous treatment}

To simulate data with a continuous treatment variable, we use the following data generating process: 

\begin{align*}
X &\sim \text{Uniform}(0,2)\\
A|X &\sim \text{Normal}(\mu=X^2-1,\sigma^2=2)\\
Y|A,X &\sim \text{Normal}(\mu = 5X + 9A(X+2)^2 + 5\text{sin}(X\pi) + 25A, \sigma^2 = 1)\\
\end{align*}

We consider a shift intervention where we replace the observed treatment $A$ with $A'$ where
\begin{align*}
    A'|A,X &\sim A + 1.
\end{align*}
The estimands of interest are the ASE and the LASE of the shift intervention where we only implement the shift intervention for individuals with observed $A < 0$. Under the DGP, the true values of the parameters are given by $\psi^{ASE} = 108.997$ and $\psi^{LASE} = 94.814$. 

At each iteration, we estimate the Riesz representer for the ASE and the LASE directly using our gradient boosting algorithm RieszBoost. For comparison, we also estimate the Riesz representer for the ASE and the LASE indirectly by first estimating the conditional density of $A$ given $X$ (via Gaussian kernel density estimation) and then plugging the estimate into the known analytical form of the Riesz representer. Details for the indirect approach can be found in Appendix \ref{appendix:est/rr_ase} and Appendix \ref{appendix:est/rr_lase}. While this approach is feasible in lower-dimensional settings, it becomes computationally expensive or impractical as the dimensionality of $X$ grows. Moreover, kernel density estimation requires selecting a parametric kernel, and it is difficult to find a straightforward nonparametric alternative. Thus, for both the ASE and LASE, RieszBoost offers an efficient and scalable alternative that bypasses these computational and practical challenges. We also estimate the outcome regression given treatment and covariate using gradient boosting regression. Finally, we compute estimates $\widehat{\psi}^{ASE}$ and $\widehat{\psi}^{LASE}$ using EEE estimators for the two parameters. 

Over 500 simulations, we draw 1,000 observations from the DGP. We use 500 as our training data set, and the other 500 serve as our validation data set. We use cross validation to train each gradient boosting algorithm over a set grid of hyperparameters. Specifically, we consider number of estimators values $M\in \{10, 30, 50, 75, 100, 150, 200\}$, learning rate values $\lambda \in \{ 0.001, 0.01, 0.1, 0.25\},$ and maximum depth of trees $\in \{3,5,7\}.$ The hyperparameter grid used was the same for the RieszBoost and GBT for outcome regression learning. To indirectly the Riesz representer, we estimate the conditional density of $A$ given $X$ using Gaussian kernel density estimation by estimating the joint density and dividing by the estimated marginal density of $X$. Gaussian kernel density estimators take in a bandwidth as a user-specified hyperparameter. We also conduct a grid search for hyperparameter tuning, selecting the hyperparameters which minimize the loss for the conditional density. We search over a grid of bandwidth $\in  \{0.01, 1.2575, 2.505 , 3.7525, 5\}$ for estimating the joint density and bandwidth $\in \{0.01, 0.5075, 1.005, 1.5025, 2\}$ for estimating the marginal density. We select the pair of bandwidths that maximizes the conditional likelihood.

\subsubsection{Results for ASE and LASE}
  
Table \ref{tab:rr_rmse_ase} compares the performance of the two estimation strategies with respect to estimating the Riesz representer in terms of RMSE and MAE over the 500 simulations. Tables \ref{tab:ase_results} and \ref{tab:lase_results} presents our main results for the ASE and LASE, respectively. While RieszBoost results in a slightly higher RMSE and MAE when estimating the Riesz representer when compared to the indirect method, both approaches achieve similar coverage for the true parameter at $\alpha = 0.05$, with RieszBoost resulting in slightly better coverage. Importantly, RieszBoost achieves these results 
\textit{without requiring the esitmation of conditional densities}, thereby avoiding the computational challenges and modeling assumptions assoicated with such estimates, particularly in high-dimensional settings.

\begin{table}[h!]
\centering
\caption{Results for Estimating $\alpha_0^{ASE}$ and $\alpha_0^{LASE}$}
\label{tab:rr_rmse_ase}
\begin{tabular}{ccccc}
                                      &                                    &                                   &                                    &                                   \\ \cline{2-5} 
\multicolumn{1}{c|}{}                 & \multicolumn{2}{c|}{\textbf{ASE}}                                      & \multicolumn{2}{c|}{\textbf{LASE}}                                     \\ \hline
\multicolumn{1}{|c|}{\textbf{Method}} & \multicolumn{1}{c|}{\textbf{RMSE}} & \multicolumn{1}{c|}{\textbf{MAE}} & \multicolumn{1}{c|}{\textbf{RMSE}} & \multicolumn{1}{c|}{\textbf{MAE}} \\ \hline
\multicolumn{1}{|c|}{RieszBoost}      & \multicolumn{1}{c|}{0.366}         & \multicolumn{1}{c|}{0.230}        & \multicolumn{1}{c|}{0.252}         & \multicolumn{1}{c|}{0.154}        \\
\multicolumn{1}{|c|}{Indirect}        & \multicolumn{1}{c|}{0.296}         & \multicolumn{1}{c|}{0.203}        & \multicolumn{1}{c|}{0.129}         & \multicolumn{1}{c|}{0.080}        \\ \hline
\end{tabular}
\end{table}

\begin{table}[!h]
\caption{Average Shift Effect Simulation Results}
\centering
\label{tab:ase_results}
\begin{tabular}{|c|c|c|c|c|c|}
\hline
\textbf{Method}     & \textbf{Avg. Estimate} & \textbf{Avg. Est. SD} & \textbf{RMSE} & \textbf{Empirical SD} & \textbf{Coverage (95\%)} \\ \hline
RieszBoost & 109.672                & 2.087                 & 2.796         & 2.713                 & 0.934                    \\
Indirect   & 109.919                & 1.985                 & 2.298         & 2.104                 & 0.928                    \\ \hline
\end{tabular}
\end{table}

\begin{table}[!h]
\caption{Local Average Shift Effect Simulation Results}
\centering
\label{tab:lase_results}
\begin{tabular}{|c|c|c|c|c|c|}
\hline
\textbf{Method}     & \textbf{Avg. Estimate} & \textbf{Avg. Est. SD} & \textbf{RMSE} & \textbf{Empirical SD} & \textbf{Coverage (95\%)} \\ \hline
RieszBoost & 94.921                 & 1.768                 & 1.859         & 1.855                 & 0.946             \\
Indirect   & 94.758                 & 1.753                 & 1.789         & 1.789                 & 0.940             \\ \hline
\end{tabular}
\end{table}

\section{Discussion}
\label{discuss}

In this paper, we introduce a new method for Riesz regression using gradient boosting called RieszBoost. This method addresses key challenges in indirectly estimating the Riesz representer via nuissance parameter estimation, providing a robust and user-friendly alternative to existing Riesz regression methods. We detail the special considerations necessary for minimizing the Riesz loss within the framework of gradient boosting. First, researchers must create the predictor matrix, which includes additional rows corresponding to pseudo-data relevant to the Riesz loss as arguments of the Riesz representer, and the target matrix, which includes additional rows corresponding variables relevant to the Riesz loss outside of arguments of the Riesz representer. This augmentation allows the algorithm to capture the effects of both the observed and counterfactual data points of interest on the loss function. Second, researchers must compute the gradient by evaluating candidate functions $\alpha$ evaluated at each row of the predictor matrix and using the values from the corresponding row of the target matrix to construct an appropriate ``residual'' function. 

We provide simulation studies showing that RieszBoost provides unbiased estimates for four causal estimands of interest, along with confidence intervals that achieve appropriate coverage. The simulations also reveal that RieszBoost performs on par with, or better than, indirect methods for estimating the Riesz representer using gradient boosting. Notably, indirect approaches to estimating the Riesz representer, which depend on preliminary estimates of nuisance parameters, can be highly variable or computationally prohibitive, especially when the regressors are high-dimensional. In contrast, gradient boosting remains computationally feasible even in high-dimensional settings. Indeed, gradient boosting has been shown, both theoretically and empirically, to perform robustly in such contexts (see, e.g., \cite{schuler_lassoed_2023}).

Gradient boosting is a powerful algorithm that works well on tabular data. Unlike neural networks, RieszBoost is relatively straightforward to train, requiring the tuning of only a small number of intuitive hyperparameters. By introducing this new algorithm for Riesz regression, we empower researchers with greater flexibility and choice in their methodological toolkit, enabling the application of direct Riesz regression methods to a broader range of problems.

\bibliographystyle{unsrt} 
\bibliography{main.bib}

\newpage
\appendix
\appendixpage
\addappheadtotoc
\section{Efficient Estimating Equations Estimators }
\label{appendix:est}

Recall that the EIF for a parameter $\Psi : \mathbb P \mapsto \mathbb E_{\mathbb P}[m(W, \mu_{\mathbb P})] = \mathbb E_{\mathbb P}[\alpha_\mathbb{P}(W)\mu_\mathbb{P}(W)] = \psi_\mathbb{P}$ is given by
$$\phi_\mathbb{P}(O) = m(W, \mu_\mathbb{P}) - \psi_\mathbb{P} + \alpha_\mathbb{P}(W)(Y-\mu_\mathbb{P}(W)).$$
Let $\hat{\alpha}$ and $\hat{\mu}$ be estimates of $\alpha_\mathbb{P}$ and $\mu_\mathbb{P}$, respectively. We can substitute these estimates into the EIF along with a placeholder estimate ($\hat\psi$) for our parameter of interest:
$$\hat{\phi}(O) = m(W, \hat{\mu}) - \hat\psi + \hat{\alpha}(W)(Y-\hat{\mu}(W)).$$
The EEE estimator $\hat{\psi}$ results from setting the empirical mean of the EIF equal to 0 and solving for $\hat\psi$ (for motivation, see \cite{schuler_introduction}):
\begin{align*}
    \hat{\psi} &= \frac{1}{n} \sum_{i=1}^n m(W_i, \hat{\mu}) + \hat{\alpha}(W_i)(Y_i-\hat{\mu}(W_i)).
\end{align*}

More generally, for parameters not of the form $\mathbb{E}_{\mathbb P}[m(W,\mu_{\mathbb P})]$, the influence function may depend on other aspects of the data-generating process. Let $\theta$ represent these other parameters (which may be scalar or function-valued), so that we can write the EIF as $\phi_\mathbb{P}(O) = \phi(O; \theta, \psi)$, in a mild abuse of notation. In the case above, we have $\theta = (\alpha, \mu)$. Given estimates  $\hat\theta$, the general EEE estimator is obtained the same way as above: $\hat\psi$ is the solution to $0=\frac{1}{n} \sum_{i=1}^n \phi(O_i; \hat\theta, \hat\psi)$.

Under general conditions, an EEE estimator has asymptotic variance $\mathbb{E}[\phi(O_i)^2]$. Thus, a consistent estimate of the sampling variance (for an estimator with $n$ observations) is given by $n^{-2}\sum_{i=1}^n \hat \phi(O_i)^2$. This is the variance estimator we use in all simulations.

Cross-fitting is typically used to meet the conditions required to ensure good asymptotic performance of EEE estimators. Let $\hat\theta$ (e.g., $\hat\mu$ and $\hat\alpha$) denote estimates learned from $\mathbf O$, a sample of $n$ observations, and let $\Acute{\mathbf O}$ denote a separate sample of $\Acute n$ observations (in practice, the two samples can be halves of one dataset). The cross-fit EEE estimate $\hat\psi$ is the solution to $0 = \frac{1}{\Acute n} \sum_{\Acute{\mathbf O}} \phi(O_i; \hat\theta, \hat\psi)$. The idea is that, in the EEE estimator, no functional components of $\hat\theta$ are ever evaluated on data used to fit them. This is analogous to using a separate test set to estimate out-of-sample error in prediction problems. The process can be generalized to multiple folds and the resulting estimates averaged to obtain a general cross-fit estimate that makes better use of the available data. In all simulations, we use a simple version of cross-fitting with $n = \Acute n$, both for the direct and indirect methods of estimating the Riesz representer.

\section{Specifics for Example Parameters}
\label{appendix:ex_params}

\subsection{Average Treatment Effect Among the Treated}

Let $W=(A, X)$, where $A$ is a binary treatment of interest and $X$ are confounders, and let $\mathbb{P}(A=1)$ be the probability of treatment over the whole population. If $A$ and $Y$ are independent given $X$ and the propensity score is bounded away from 1, then the average treatment effect among the treated (ATT) can be written 
    \begin{align*}
    \Psi(\mathbb{P}_0) &= E\left[\left(
    \mathbb{E}[Y|A=1, X]-\mathbb{E}[Y|A=0, X]\right)\big|A=1
    \right]\\
    &= \frac{1}{\mathbb{P}(A=1)} \mathbb E\left[A\big(\mu_0(1,X) - \mu_0(0,X)\big)\right].
    \end{align*}

\label{appendix:est/rr_att}

   As shown in \cite{hubbard_direct_2011}, the EIF of the ATT is given by
    \begin{equation}
    \label{appendix:att_eif}
\phi_0(O) = \frac{1}{\mathbb{P}(A=1)} \left[A\big(\mu_0(1,X) - \mu_0(0,X) - \Psi(\mathbb{P}_0) \big) + \underbrace{\left(A - \frac{(1-A)\pi_0(X)}{1-\pi_0(X)}\right)}_{\alpha_0(A,X)} (Y-\mu_0(A,X)) \right]
\end{equation}
where $\pi_0(x) = \mathbb{P}(A=1|X=x)$ is the propensity score. Note that this EIF does not take the same form as in equation \ref{gen_eif}. This is because our parameter depends on the true distribution not only through the regression function $\mu_0$, but also through the probability $\mathbb{P}(A=1)$. Therefore, to derive the EIF, we must take this dependency into consideration (e.g. using the delta method on the inverse probability parameter $\frac{1}{\mathbb{P}(A=1)}$ and the ``partial'' parameter $E\left[A\big(\mu(1,X) - \mu(0,X)\big)
\right]$). The partial parameter does satisfy the form $\mathbb{E}[m(O,\mu)]$, and its Riesz representer, which is $\alpha_0(A,X) = A - \frac{(1-A)\pi_0(X)}{1-\pi_0(X)}$, shows up in the EIF for the full parameter. Thus, there is still a use for Riesz regression.

To construct an EEE estimator of this parameter we need estimates of $\mu_0$ (via standard regression), $\alpha_0$ (via either Riesz regression or an indirect approach), and $\mathbb{P}(A=1)$ (which we estimate with a sample mean). 

\subsubsection{Indirect Riesz Representation Estimation}

For our benchmark indirect estimator, we estimate $\pi_0(x)$ using a classifier and substitute this estimator into $\alpha_0$. Letting $\hat\pi(x)$ be our estimate of the propensity score, our indirect estimate of $\alpha_0$ is given by

$$\hat\alpha^{\text{indirect}}(a,x) = a - \frac{(1-a)\hat\pi(x)}{1-\hat\pi(x)}$$

\subsubsection{RieszBoost}
\label{appendix:att_rb}

    As far as Riesz regression is concerned, we will focus on the ``partial'' parameter $\mathbb{E}[A\big(\mu(1,X) - \mu(0,X)\big)]$. The linear functional of interest is given by $m(O,\mu) = A\big(\mu(1,X) - \mu(0,X)\big)$.
    The empirical Riesz loss is therefore
    $$L_n(\alpha) = \frac{1}{n} \sum_{i=1}^n -2A_i(\alpha(1, X_i) - \alpha(0, X_i)) + \alpha(A_i, X_i)^2.$$
    
    Taking derivatives, we compute the empirical gradient (up to a multiplicative constant):
   $$
   \nabla_{\alpha}L_n(a,x) = \sum_{i=1}^n 1\big((a,x) = (A_i,X_i)\big)\alpha(a,x) - 1(x =X_i)A_i\big(1(a=1)-1(a=0)\big).
   $$
   Let $\mathbf{X}^1 = \{X_i: A_i = 1\}_1^n$ be the set of observed $X_i$ for treated individuals and $\mathbf{X}^0 = \{X_i: A_i = 0\}_1^n$ be the set of observed $X_i$ for untreated individuals. Suppose we have $n_1$ treated individuals and $n_0$ untreated individuals. The empirical gradient is only non-zero for points 
   $\{
   (x,a): 
   x \in \mathbf X, 
   a \in \mathbf A  
   \} \cup \{
   (x,a): 
   x \in \mathbf{X}^1, 
   a = 0  
   \}$.  Therefore, we can construct the predictor matrix with $n+n_1$ rows:
   $$
    \tilde{\mathbf W} = 
    \begin{blockarray}{cc}
    \tilde{A} & \tilde{X} \\
    \begin{block}{[cc]}
    \mathbf 1 & \mathbf X^1\\
    \mathbf 0 & \mathbf X^0\\
    \mathbf{0} & \mathbf X^1 \\
    \end{block}
    \end{blockarray}.
    $$
    
    \noindent We also see that the empirical gradient at each row in the predictor matrix depends on the original and counterfactual treatment values. Therefore, we can construct the target matrix:
    
    $$
    \tilde {\mathbf Z} = 
    \begin{blockarray}{cc}
    \tilde{A} & A^\degree \\
    \begin{block}{[cc]}
    \mathbf 1 & \mathbf 1\\
    \mathbf 0 & \mathbf 0\\
    \mathbf{0} & \mathbf 1 \\
    \end{block}
    \end{blockarray}.
    $$

    Finally, we can define the residual function where, for each $j = 1, \dots, n+n_1$, we evaluate

    $$r(\tilde{Z}_j, \hat{Z}_j) = -1(\tilde{A}_j=A^\degree_j)\hat Z_j  + 1(A^\degree_j=1)\big(2\tilde{A}_j - 1\big),$$

    \noindent where $\tilde{Z}_j = \hat{\alpha}(\tilde A_j, \tilde X_j)$ and $\hat{Z}_j = (\tilde A_j, A^\degree_j)$. Riesz regression for the partial ATT can be accomplished by using the predictor matrix, target matrix, and residual function in Algorithm \ref{algo:rieszboost}.

\label{appendix:est/att}

\subsection{Average Shift Effect}
In the main text we describe the ASE, give its EIF, and show the form of its Riesz representer. To construct an EEE estimator for the ASE, we only require estimates of the Riesz representer and outcome regression.

\subsubsection{Indirect Riesz Representation Estimation}
\label{appendix:est/rr_ase}
For our benchmark indirect estimator, we estimate $p_{A|X}(A,X)$ using Gaussian kernel density estimation and plug this estimate into $\alpha_0$. We first estimate the joint density of $A$ and $X$ and the marginal density of $X$, and then take the ratio. Let $\widehat{p_{A|X}}(a,x)$ be our estimate of $p_{A|X}(A,X)$. Then, our estimate of $\alpha_0$ is given by
$$\hat\alpha^{\text{indirect}}(a,x) = \frac{\widehat{p_{A|X}}(a-\delta,x)}{\widehat{p_{A|X}}(a,x)} - 1.$$

\subsection{Local Average Shift Effect}
 Let $W=(A,X)$, where $A$ is a continuous exposure of interest and X are confounders, $p_{A|X}(A,X)$ is the true conditional density of treatment given confounders, and let $\mathbb{P}(A<t)$ be the probability of treatment being less than some value $t$ over the whole population. We are interested in estimating the local average shift effect (ASE), a generalized ATT involving shift interventions, described in \cite{susmann_longitudinal_2024}. We consider an additive increase in treatment  of $\delta$ (covariate values held constant) among individuals with a treatment value below a set threshold $t$. Our parameter of interest is the expected outcome under such policy relative to the average observed outcome among those who experience the increase:
 \begin{align*}
     \Psi(\mathbb{P}_0)^{ASE} &= \mathbb{E}[\mathbb{E}[Y|A+\delta, X]-\mathbb{E}[Y|A, X]|A<t]\\
     &= \frac{1}{\mathbb{P}(A<t)}\mathbb{E}[1(A<t) \big(\mu_0(A+\delta, X) - \mu_0(A,X) \big)].
 \end{align*}
Under identification assumptions outlined in \cite{susmann_longitudinal_2024}, this parameter captures the causal effect of increasing treatment by $\delta$ units across the subpopulation of individuals with treatment values $A<t$.

The EIF of the LASE is given by
\begin{equation}
\label{appendix:lase:eif}
\phi_0(O) = \frac{1}{\mathbb{P}(A<t)} \Big[1(A<t)\big(\mu_0(A+\delta,X) - \mu_0(A,X) - \Psi(\mathbb{P}_0) \big) + \alpha_0(A,X) \big(Y-\mu_0(A,X)\big)\Big],
\end{equation}
where $\alpha_0 = 1(A<t+\delta)\left(\frac{p_{A|X}(A-\delta,X)}{p_{A|X}(A,X)}\right) - 1(A<t).$

 Note that, similar to the ATT example in Appendix \ref{appendix:att_rb}, this EIF does not take the same form as in equation \ref{gen_eif}. 
 Since our parameter depends on the true distribution not only through the regression function $\mu_0$, but also through the probability $\mathbb{P}(A<t)$, we must take this into account when calculating the EIF. To derive the EIF, we must use the delta method on the inverse probability parameter $\frac{1}{\mathbb{P}(A<t)}$ and the ``partial'' parameter $E\left[1(A<t)\big(\mu(A+\delta,X) - \mu(A,X)\big)\right]$. The partial parameter does have the form $\mathbb{E}[m(O, \mu)]$, and its Riesz representer $\alpha_0$ appears in the EIF for the full parameter. Thus, there is still a use for Riesz regression. 

To construct an EEE estimator of this parameter, we need estimates of $\mu_0$ (via standard regression), $\alpha_0$ (via either Riesz regression or an indirect approach), and $\mathbb{P}(A<t)$ (which we estimate with a sample mean). 

\subsubsection{Indirect Riesz Representation Estimation}
\label{appendix:est/rr_lase}
For our benchmark indirect estimator, we estimate $p_{A|X}(a,x)$ using Gaussian kernel density estimation and plug this estimate into $\alpha_0$. We first estimate the joint density of $A$ and $X$ and the marginal density of $X$, and then take the ratio. Let $\widehat{p_{A|X}}(a,x)$ be our estimate of $p_{A|X}(a,x)$. Then, our estimate of $\alpha_0$ is given by
$$\hat\alpha^{\text{indirect}}(a,x) = 1(a<t+\delta)\left(\frac{\widehat{p_{A|X}}(a-\delta,x)}{\widehat{p_{A|X}}(a,x)}\right) - 1(a<t)$$

\subsubsection{RieszBoost}
For Riesz regression, we will focus on the partial parameter $E\left[1(A<t)\big(\mu_0(A+\delta, X) - \mu_0(A, X)\big)\right]$. The linear functional of interest is given by $m(O,\mu) = 1(A<t)\big(\mu(A+\delta, X) - \mu(A, X)\big)$. The empirical Riesz loss is therefore
    $$L_n(\alpha) = \frac{1}{n} \sum_{i=1}^n -2 \left[1(A_i<t)\big(\alpha(A_i+\delta, X_i) - \alpha(A_i, X_i)\big)\right] + \alpha(A_i, X_i)^2.$$
And, taking derivatives, we compute the empirical gradient (up to a multiplicative constant)
$$\nabla_{\alpha}L_n(a,x) = \sum_{i=1}^n 1\big((a,x) = (A_i,X_i)\big)\alpha(a,x) - 1(x =X_i)1(A_i<t)\big(1(a=A_i+\delta)-1(a=A_i)\big).$$
   Let $\mathbf{X}^1 = \{X_i: A_i < t\}_1^n$ be the set of observed $X_i$ for individuals with $A_i < t$ (those who experience the increase), $\mathbf{X}^0 = \{X_i: A_i \geq t\}_1^n$ be the set of observed $X_i$ for individuals with $A_i \geq t$ (those who do not experience the increase), $\mathbf{A}^1$ be the set of observed $A_i$ for individuals who experience the treatment increase, and $\mathbf{A}^0$ be the set of observed $A_i$ for individuals who do not experience the treatment increase.
   Suppose we have $n_1$ treated individuals and $n_0$ untreated individuals. This empirical gradient is only non-zero for arguments 
   $\{
   (x,a): 
   x \in \mathbf X, 
   a \in \mathbf A  
   \} \cup \{
   (x,a): 
   x \in \mathbf{X}^1, 
   a =   \mathbf{A}^1+\delta
   \}$.
   Therefore, we can construct the predictor matrix with $n+n_1$ rows:
   $$
    \tilde{\mathbf W} = 
    \begin{blockarray}{cc}
    \tilde{A} & \tilde{X} \\
    \begin{block}{[cc]}
    \mathbf A^1 & \mathbf X^1\\
    \mathbf A^0 & \mathbf X^0\\
    \mathbf A^1+\delta & \mathbf X^1 \\
    \end{block}
    \end{blockarray}.
    $$
    
    \noindent We also see that the empirical gradient at each row in the predictor matrix depends on the original and counterfactual treatment values. Therefore, we can construct the target matrix:
    
    $$
    \tilde {\mathbf Z} = 
    \begin{blockarray}{cc}
    \tilde{A} & A^\degree \\
    \begin{block}{[cc]}
    \mathbf A^1 & \mathbf A^1\\
    \mathbf A^0 & \mathbf A^0\\
    \mathbf A^1+\delta & \mathbf A^1 \\
    \end{block}
    \end{blockarray}.
    $$

    Finally, we can define the residual function where, for each $j = 1, \dots, n+n_1$, we evaluate

    $$r(\tilde{Z}_j, \hat{Z}_j) = -1(\tilde{A}_j=A^\degree_j)\hat Z_j  + 1(A^\degree_j<t)\big(1(\tilde{A}_j\neq A_j^\degree)-1(\tilde{A}_j=A_j^\degree)\big),$$

    \noindent where $\tilde{Z}_j = \hat{\alpha}(\tilde A_j, \tilde X_j)$ and $\hat{Z}_j = (\tilde A_j, A^\degree_j)$. 
  Riesz regression for the partial parameter for the ASE can be accomplished by using the predictor matrix, target matrix, and residual function in Algorithm \ref{algo:rieszboost}.

\end{document}